\documentclass[10pt,twocolumn,letterpaper]{article}

\usepackage{cvpr}
\usepackage{times}
\usepackage{epsfig}
\usepackage{graphicx}
\usepackage{amsmath}
\usepackage{amssymb}

\usepackage[breaklinks=true,bookmarks=false]{hyperref}

\cvprfinalcopy 

\def\cvprPaperID{****} 

\begin{document}

\title{\normalsize \textbf{3D Based Landmark Tracker Using Superpixels Based Segmentation for Neuroscience and Biomechanics Studies}}
\author{Omid Haji Maghsoudi\\
Spence Lab., Bioengineering, \\
Temple University, \\
Philadelphia, PA, USA, 19122\\
{\tt\small o.maghsoudi@temple.edu}
\and
Andrew Spence$^{1}$\\
Spence Lab., Bioengineering, \\
Temple University, \\
Philadelphia, PA, USA, 19122\\
{\tt\small http://www.spencelab.com}
}

\maketitle
\footnotetext[1]{This material is based upon work supported by, or in part by, the U.S. Army Research Laboratory and the U. S. Army Research Office under contract/grant number W911NF1410141, proposal 64929EG, to A. Spence.}

\begin{abstract}
Examining locomotion has improved our basic understanding of motor control and aided in treating motor impairment. Mice and rats are premier models of human disease and increasingly the model systems of choice for basic neuroscience. High frame rates (250 Hz) are needed to quantify the kinematics of these running rodents. Manual tracking, especially for multiple markers, becomes time-consuming and impossible for large sample sizes. Therefore, the need for automatic segmentation of these markers has grown in recent years. Here, we address this need by presenting a method to segment the markers using the SLIC superpixel method. The 2D coordinates on the image plane are projected to a 3D domain using direct linear transform (DLT) and a 3D Kalman filter has been used to predict the position of markers based on the speed and position of markers from the previous frames. Finally, a probabilistic function is used to find the best match among superpixels. The method is evaluated for different difficulties for tracking of the markers and it achieves 95\% correct labeling of markers.
\end{abstract}

\section{Introduction}
Studying animal, including humans, locomotion has been one of the challenging areas in the modern science. Our health and well-being are directly linked with movement. Animal movement can explain some biological world phenomena. In addition, it can impact the treatment of musculoskeletal injuries and neurological disorders, improve prosthetic limb design, and aid in the construction of legged robots \cite{maghsoudi2015novel}. 

The intentional changes in an animal gait, the timing of paw motion relative to each other using by the animal \cite{deumens2007catwalk}, can be seen during movement. The animal movement can be perturbed using an internal or external perturbation. A mechanical perturbation (e.g., earthquake) while the animal is running, for example, deflecting the surface during running, or an electrical stimulation applied to the nervous system, or even the application of new genetically targeted techniques, like optogenetics \cite{deisseroth2011optogenetics} or designer receptors exclusively activated by designer drugs \cite{roth2016dreadds}, are several of the increasingly sophisticated methods applying perturbations that dissect the movement control.

To study the gait and kinematics, it is needed to track specific landmarks on the body of an animal. Tracking of these landmarks on the body of animal relies on shaving fur, drawing markers on the skin, attaching retroreflective markers, or manual clicking by a user for consecutive frames{\cite{maghsoudi2016rodent}}. The attachment of retroreflective markers can be impossible in many cases as animals, like rats and mice, start grooming and chewing the markers. Therefore, shaving fur and drawing markers can be the most reliable method for tracking of specific landmarks \cite{schubert2015automatic}.

Commercially available systems (Digigait \cite{dorman2014comparison, gadalla2014gait, nori2015long}, Motorater \cite{preisig2016high}, Noldus Catwalk \cite{deumens2007catwalk, hamers2001automated, huehnchen2013assessment, parvathy2013gait}) are prohibitively expensive, and may only provide information about paws during the stance phase which makes them limited for some studies. In addition, some computerized methods (simple thresholding, cross-correlation, or template matching) have been proposed to answer this need. However, manual clicking can be considered the usual method to track the markers \cite{hedrick2008software}. Therefore, the need for a robust method to help neuroscientists and biologists have been felt.

Tracking has been a recently popular topic in image processing. Many methods have been developed for different applications; cell migration tracking \cite{penjweini2017investigating}, human tracking \cite{ma2016counting}, and diseases frames tracking in consecutive frames \cite{haji2012automatic, mahdi2017detection}. Tracking methods should be developed based on conditions governing around a specific problem which make them unique \cite{Maghsoudi17IET}.

As discussed, locomotion analysis needs to track some landmarks on the body of an animal. The 2D tracking from the frames can provide the required knowledge to examine the animal's gait. However, access to 3D information can improve our understanding about locomotion including roll, pitch, and yaw \cite{migliaccio2011characterization}.

We presented a superpixel based segmentation method to find the markers following by a weighted 2D tracker \cite{maghsoudi2017superpixels}. The results were so promising and it inspired us to use the SLIC method for segmentation. The tracker was using 2D information from an image plane to find the position of landmarks for consecutive frames. However, the latest issue caused problems for tracking of landmarks when they were occluding by the body or another limb, getting too close to another landmark, or having some dirt on the plexy glass.

Here we take the advantage of superpixels for segmentation of landmarks, with a small difference compared to \cite{maghsoudi2017superpixels}. However, the main contribution of this study is a robust tracker design to resolve the limitations of our previously presented method. We find the solution in using of 3D information and processing two camera information at the same time. 3D Kalman filter and direct linear transform were used to achieve this goal.

\begin{figure*}[t!p]
\centering
\includegraphics[width=0.7\textwidth]{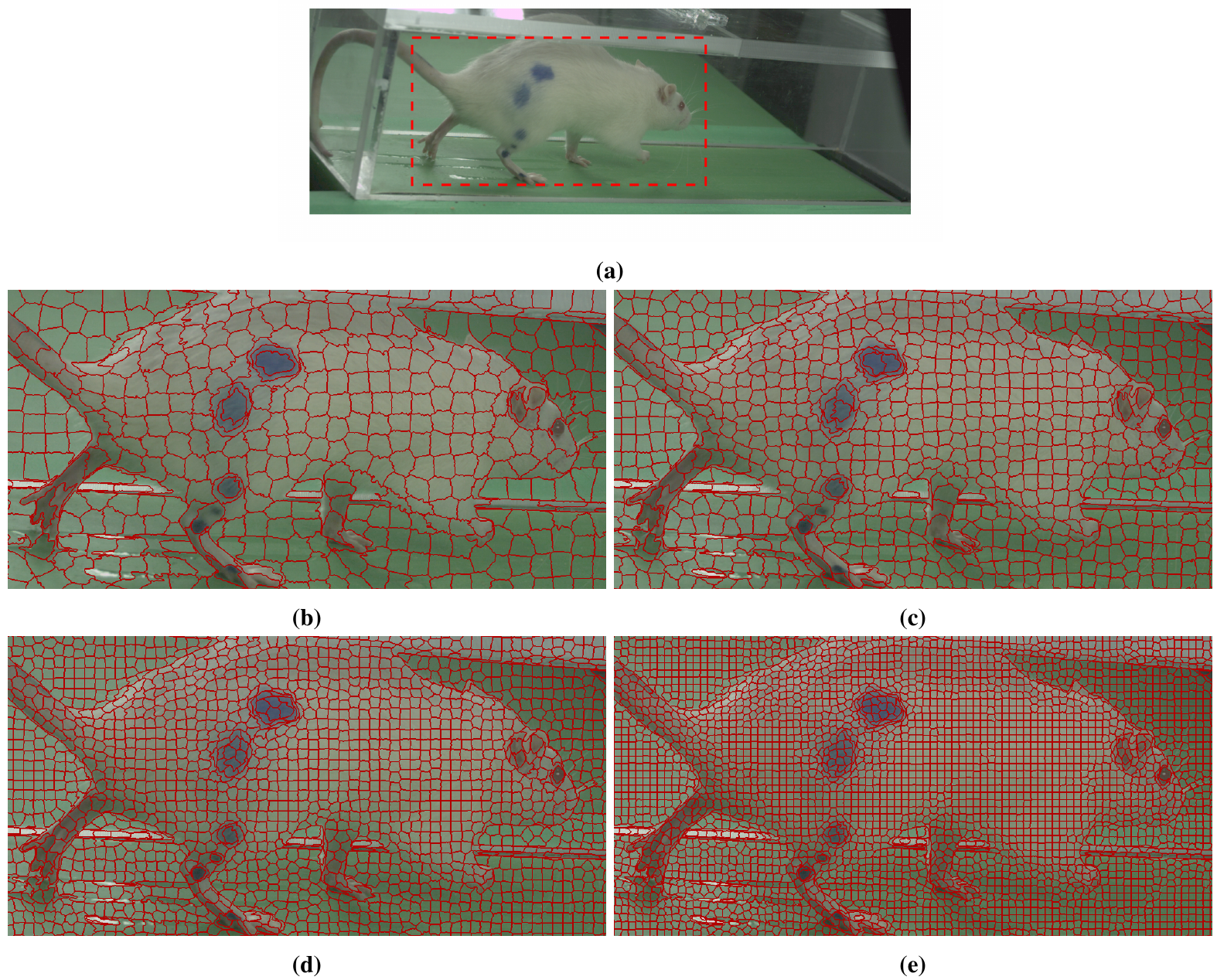}
\caption{A sample video frame of rat locomotion with five markers drawn on the right side of an animal.  A shows the original frame with the drawn red rectangle showing a region of a frame which is zoomed in for a better visualization. B, c, d, and e illustrate the zoomed in area from image (a) with 1250, 2500, 5000, and 10000 superpixels.} \label{fig1}
\end{figure*}

\section{Methods}
\subsection{Camera and Treadmill Setup}
Four side view cameras were used to capture video from a treadmill located in the middle of capturing area. The cameras were synchronized using an external pulse generator inducing a 250 HZ pulse to assure they were captured at the same time. In addition, the frames were labeled by a UTC time provided the pulse generator to not miss a single frame. The capture time for each trial was 4 seconds providing 1000 frames. The frames were Bayer encoded and we use a debayering function to convert them to RGB color space frames \cite{maghsoudi2016rodent}. 

We converted the frames from the RGB color space to the HSV color space because it places all color information in a single channel, as compared to the RGB or the LAB colors spaces \cite{hajimaghsoudi2012automatic, Maghsoudi16_2}. 

\subsection{Superpixel Segmentation} \label{SLIC}
Superpixels contract and group uniform pixels in an image and have been widely used in many computer vision applications such as image segmentation \cite{Li12, Mori04}. the outcome is more natural and perceptually meaningful representation of the input image compared to pixels. Different approaches have been developed to generate superpixels: normalized cuts \cite{Ren03}, mean shift algorithm \cite{Comaniciu02}, graph-based method \cite{Felzenszwalb04}, Turbopixels \cite{Levinshtein09}, SLIC superpixels \cite{Achanta12}, and optimization-based superpixels \cite{Veksler10}. Simple linear iterative clustering (SLIC) \cite{Achanta12} generates superpixels relatively faster than other methods.

SLIC speed performance depends on a number of superpixels and the size of an image. Considering the size of image constant, the number of superpixels plays as the key parameter. Having $N$ superpixels divides the image to $N$ initial squares and associate the center of each square as the cluster center. This center should not be on an edge of an object; therefore, the center is transferred to the lowest gradient position in a $3\times3$ neighborhood. Based on color information of each pixel with its nearest cluster centers, the pixel would be associated with a cluster center.  It means that two coordinate components ($x$ and $y$) depict the location of the segment and three components (for example in the RGB color space, $R$, $G$, and $B$) are derived from color channels. SLIC calculates a distance (an Euclidean norm on 5D spaces) function, which is defined as follow, and try to match the pixels based on this function.
\begin{equation}
\label{eq:1}
D_{c} = \sqrt{(R_{j}-R_{i})^{2}+(G_{j}-G_{i})^{2}+(B_{j}-B_{i})^{2}},
\end{equation}
\begin{equation}
\label{eq:2}
D_{p} = \sqrt{(x_{j}-x_{i})^{2}+(y_{j}-y_{i})^{2}},
\end{equation}
\begin{equation}
\label{eq:3}
D = \sqrt{(\frac{D_{c}}{N_{c}})^{2}+(\frac{D_{p}}{N_{p}})^{2}}.
\end{equation}
where $N_{c}$ and $N_{p}$ are respectively maximum distances within a cluster used to normalized the color and spatial proximity. SLIC calculates this function for the cluster centers located in twice width of the initial square to minimize the calculation process.

The results of superpixel segmentation with four different SLIC superpixel numbers on a sample frame captured from a rat by five markers supposed to be drawn on the body is illustrated in Figure \ref{fig1}. Although, six markers were drawn because of a human error in the drawing. This frame was intentionally selected to show that a high number of superpixels can help us to segment the small markers and fix the human error for drawing the markers. The human error like this would not affect the segmentation and tracking; however, it can affect the tracking if they, the mistake and real marker, would be too close to each other (less than 10 pixels) which makes it impossible to differentiate them in some cases. Fortunately, a mistake like this is rare.

\subsection{Direct Linear Transform} \label{DLT}
Direct linear transform (DLT) has been proposed to calibrate cameras for generating 3D reconstruction from the captured frames \cite{hatze1988high, pvribyl2017absolute}. It has been used to create a 3D model of objects in different applications, especially in biology and biomechanics worlds \cite{choo2003improved, hedrick2008software, hedrick2012morphological, theriault2014protocol, song2014three}. 

Figure \ref{fig2} shows how an object can be projected to the camera image plane. $O$ with $[x,y,z]$ is the an object in 3D space. $N1$ with $[x_{1},y_{1},z_{1}]$ and $N2$ with $[x_{2},y_{2},z_{2}]$ are the camera projection point which they project the object to $I1$ with $[U1,V1]$ coordinate in the image plane of camera 1 (the $U1V1$ space) and $I2$ with $[U2,V2]$ coordinate in the image plane of camera 2 (the $U2V2$ space). DLT gives the following relation between the object coordinate and projected object on the image plane from camera 1:
\begin{equation}
U1 = \frac{L1X+L2Y+L3Z+L4}{L9X+L10Y+L11Z+1}
\end{equation}
\begin{equation}
U2 = \frac{L5X+L6Y+L7Z+L8}{L9X+L10Y+L11Z+1}
\end{equation}

To find $L1$ to $L11$, it is needed to calibrate the cameras. Camera calibration should be done using a calibration object having some specific markers with known coordinates. We used a custom-made Lego with attached balls on top. This Lego can be seen in Figure \ref{fig3}.

\subsection{3D Kalman Filter} \label{Kalman}
Kalman filter for motion analysis uses some observed measurements over time and estimates variables related to the motion.  Kalman filter have been used frequently to predict the position of objects in different fields, human tracking \cite{ligorio2015novel}, mice tracking \cite{Spence13}, or cardiovascular disease detection \cite{bersvendsen2016automated}. The Kalman filter model assumes that the state of a system for a frame n evolved from the prior state at frame n-1 as follow \cite{kalman1960new}:
\begin{equation}
x{n} = A_{n}x_{n-1}^{2}+B_{n}x_{n-1}+C_{n}E_{n}+D_{n}
\end{equation}
where $x$, $E$, and $t$ are the position of object, external force causing changes in position, and frame number. $A$, $B$, $C$, and $D$ are four coefficients for each frame. We considered that there is no external and acceleration causing changes in our system to simplify the system. Considering having three dimensions, we got the following equations:
\begin{equation}
\begin{split}
X{n} = B1_{n}X_{n-1}+D1_{n} \\
Y{n} = B2_{n}Y_{n-1}+D2_{n} \\
Z{n} = B3_{n}Z_{n-1}+D3_{n}
\end{split}
\end{equation}
where $X$, $Y$, $Z$ are the coordinates in the object space seen in Figure \ref{fig2}. Therefore, our system had three states and three measurements to update the coefficients.

\subsection{Features} \label{Features}
\begin{figure}[t!p]
\centering
\includegraphics[width=0.45\textwidth]{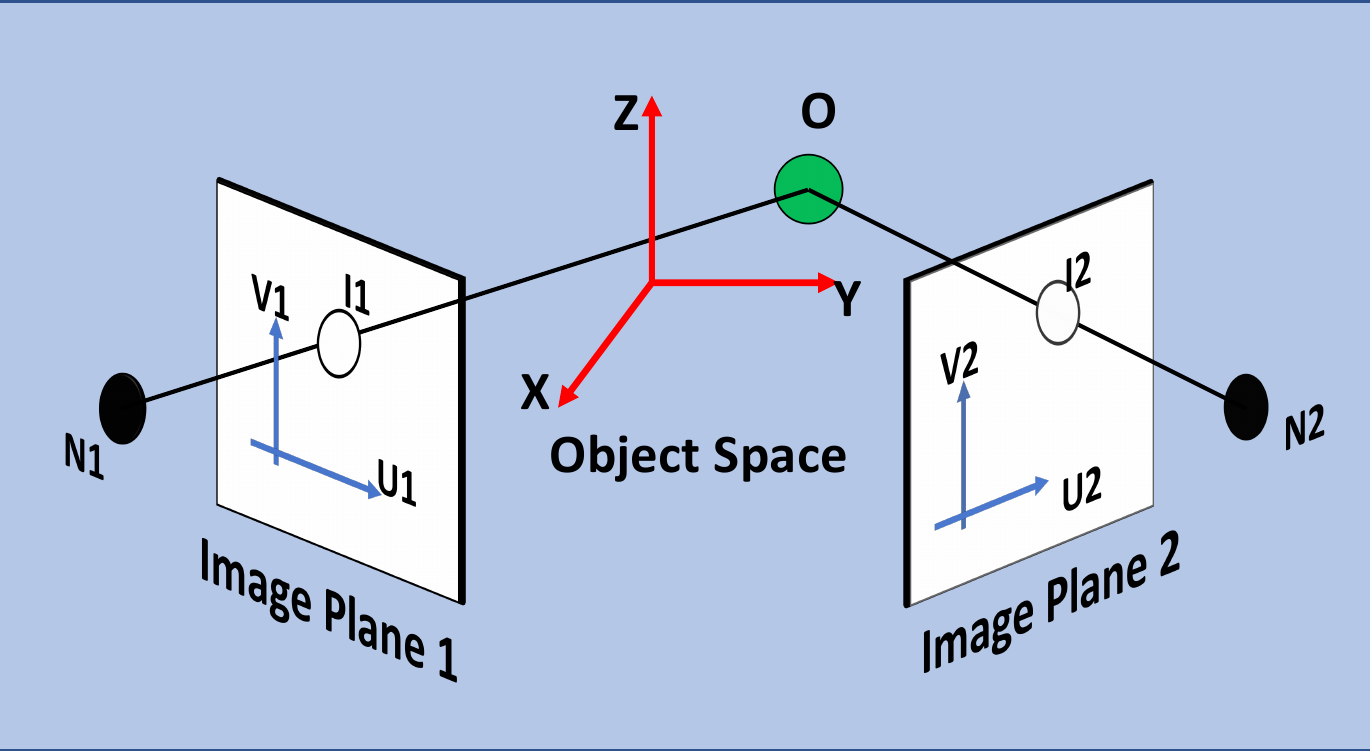}
\caption{This figure shows how two camera can be used to extarct the DLT coefficients from those two cameras image plane. $O$ is the object located in a 3D object space with $X$, $Y$, and $Z$ coordinate system. $N1$ and $N2$ are the projection point in camera 1 and 2, respectively. $I1$ and $I2$ are the projected points of object in image plane 1 and 2, respectively.} \label{fig2}
\end{figure}

\begin{figure}[t!p]
\centering
\includegraphics[width=0.48\textwidth]{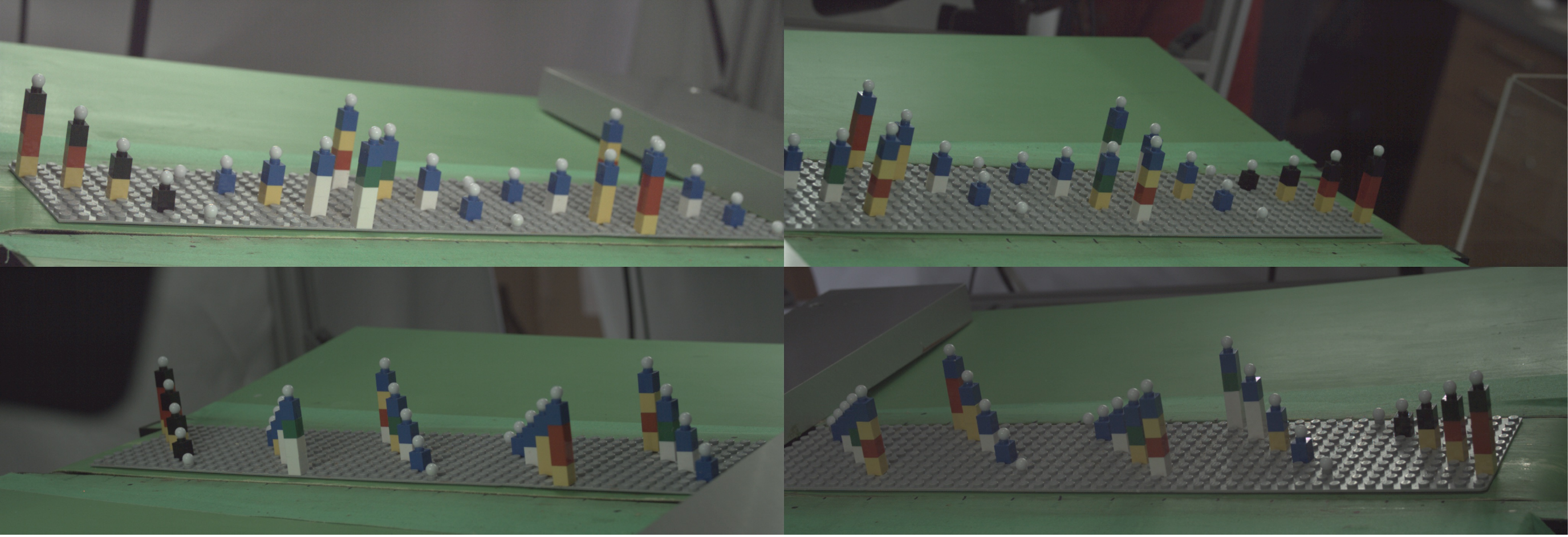}
\caption{The calibration objects used for extracting the DLT coefficients. The calibration object had 25 balls located at different heights and locations with known coordinates relative to one of the balls, the one located on a corner with the lowest height.} \label{fig3}
\end{figure}

Seven features were extracted from each superpixel to find an object having the best match with the previous detected landmarks. These seven features can be formalized as follow:
\begin{flalign}
&F_{1[i,m,n]} = Mean(S_{SP_{[i,n]}}) - Mean(S_{D_{[m,n-1]}}) \nonumber \\
&F_{2[i,m,n]} = Mean(S_{SP_{[i,n]}}) - Mean(S_{D_{[m,0]}}) \nonumber \\
&F_{3[i,m,n]} = Mean(H_{SP_{[i,n]}}) - Mean(H_{D_{[m,n-1]}}) \nonumber \\
&F_{4[i,m,n]} = Mean(H_{SP_{[i,n]}}) - Mean(H_{D_{[m,0]}}) \nonumber \\
&F_{5[i,m,n]} = Mean(G_{SP_{[i,n]}}) - Mean(G_{D_{[m,n-1]}}) \nonumber \\
&F_{6[i,m,n]} = Mean(G_{SP_{[i,n]}}) - Mean(G_{D_{[m,0]}}) \nonumber \\
&F_{7[i,m,n]} = sqrt([Mean(U_{SP_{[i,n]}}) - Mean(U_{P_{[m,n]}})]^{2} +  \nonumber \\
&\ \ \ \ \ \ \ \ \ \ \ \ \ \ \ \ \ \  [Mean(V_{SP_{[i,n]}}) - Mean(V_{P_{[m,n]}})]^{2})
\label{eq8}
\end{flalign}
where $i$, $m$, and $n$ are the superpixel number (for all superpixels in a sub-image), the marker number (five markers), and the frame number (1000 frames in our studies for each trial). $SP_{[i,n]}$, $D_{[m,n]}$, and $P_{[m,n]}$ are the superpixel number $i$ for the frame $n$, the detected landmark number $m$ for frame number $n$, and the predicted position of lanmark number $m$ for frame number $n$. $F_{1[i, m, n]}$ to $F_{7[i, m, n]}$ are the seven features corresponding to $SP_{[i,n]}$. In addition, $S$, $H$, $G$, $U$, and $V$ are saturation channel from the HSV color space \cite{hajimaghsoudi2012automatic}, hue channel from the HSV color space \cite{Maghsoudi16_2}, gray scale intensity, horizontal coordinate in the image plane (Figure \ref{fig2}), and vertical coordinate in the image plane (Figure \ref{fig2}), respectively. 

\subsection{Initial Tracker} \label{ITracker}
The initial tracker was a simple step but necessary to initialize the marker position for two consecutive frames. Two frames from each camera were needed to update the Kalman filter coefficients as speed was needed to be calculated. The frames captured from every two Cameras located on the same side were processed at the same time to have a DLT based 3D reconstruction model. 

The Initial tracker generated superpixels, by an initial value for the number of superpixels, asking a user to zoom in for a better resolution and click on the five landmarks for the first frame from each camera. Then, the maximum and minimum of all five landmarks coordinates on an image plane were calculated. This maximum and minimum numbers in each direction ($U$ and $V$) were added by 100 to make sure that not missing the marker for the next frame. A smaller sub-image was extracted to reduce the required time for performing SLIC superpixels method. The features described in section \ref{Features} would be extracted and created the initial values needed in equation \ref{eq8}. Finally, the user was asked to click the markers for the second time. The rest of frames were processed using the method described in section \ref{GTracker}.

\begin{figure*}[t!p]
\centering
\includegraphics[width=0.8\textwidth]{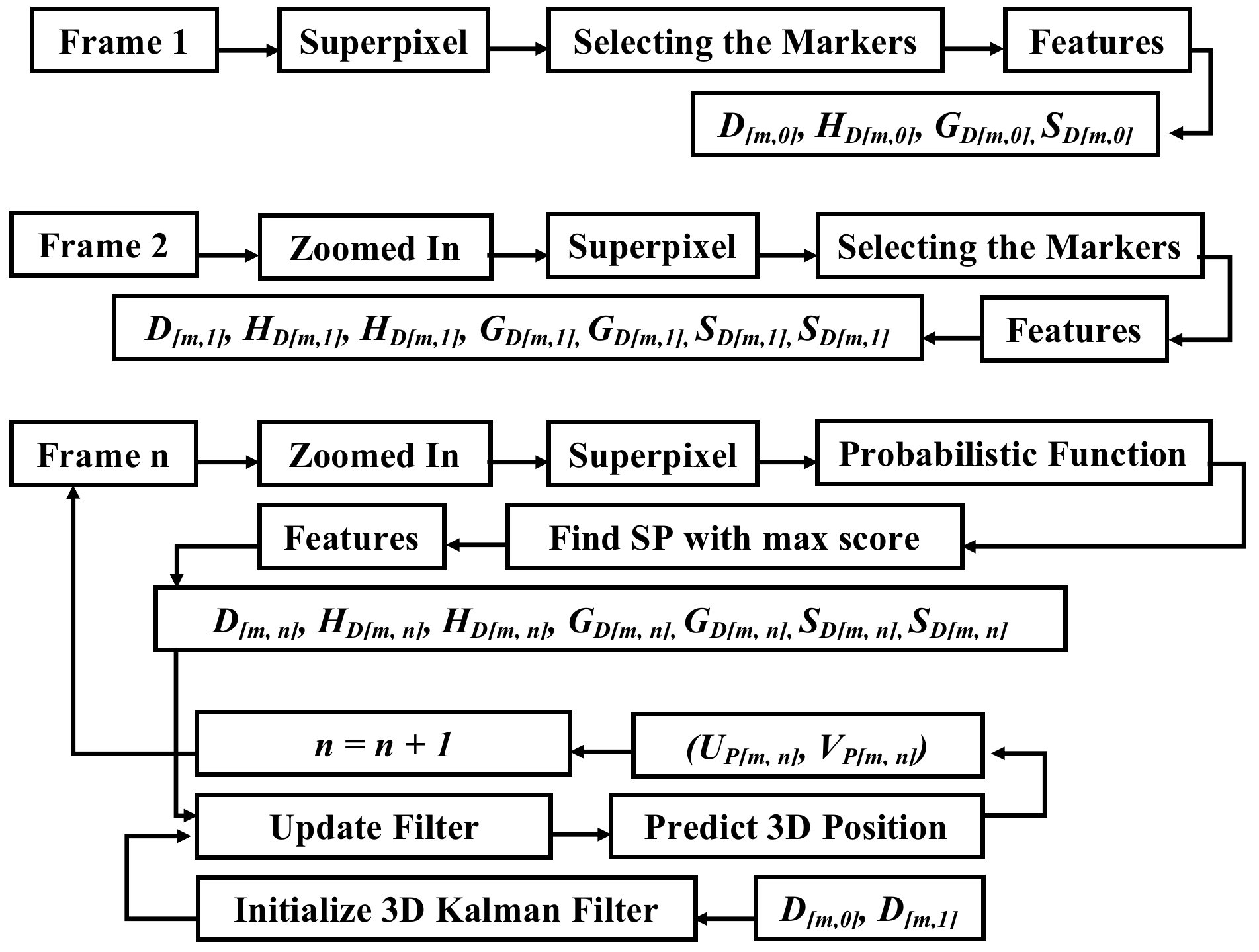}
\caption{The calibration objects used for extracting the DLT coefficients. The calibration object had 25 balls located at different heights and locations with known coordinates relative to one of the balls, the one located on a corner with the lowest height.} \label{fig4}
\end{figure*}

\subsection{General Tracker} \label{GTracker}
We subsequently focused on a $100\times100$ pixel region of interest (ROI) given by the 2D projection of the 3D coordinate predicted using 3D Kalman filter described in section \ref{Kalman}. This point has 2D coordinates of [$U_{P_{[m,n]}}$, $V_{P_{[m,n]}}$]  in the image plane for the marker number m and frame number n which it was used in equation \ref{eq8} for calculation of $F_{7[i, m, n]}$. This zoomed in the region is referred as sub-image. The size of the image was selected based on the maximum displacement of the center of the body in rats (50 pixels); the same number can be applied to mice too. 

By applying the SLIC method on each of these sub-images, the features described in section \ref{Features} were extracted for each of superpixels. One of the options for having these features was using a classifier like support vector or neural network as we presented a method for mice paw tracking using thresholding segmentation and both classifiers \cite{Maghsoudi17IET}. However, it should be reminded that the markers might have different intensities and even colors which makes the usage of a classifier limited as a tracker. 

Therefore, we developed a probabilistic function, inspired by the one we presented in \cite{Maghsoudi17IET}, to help us for tracking. First, to normalize the features and have the probabilistic function, we subtracted the $min(F_{k[\forall i, m, n]})$ from the data and divided the data by the range ($max(F_{k[\forall i, m, n]})$ - $min(F_{k[\forall i, m, n]})$) where $k$, $max$, $min$ are the feature number, maximum function, and minimum function, respectively. Therefore, the normalized feature ($N$) can be written as follow:
\begin{equation}
N_{k[i, m, n]} = \frac{F_{k[i, m, n]} - min(F_{k[\forall i, m, n]})}{max(F_{k[\forall i, m, n]}) - min(F_{k[\forall i, m, n]})}
\end{equation}

The normalized features are weighted based on the importance of features using the following array:
\begin{equation}
W = [3,1,3,2,2,1,3]
\end{equation}

Then,  each of $F_{k[\forall i, m, n]}$ should be multiplied by the corresponding $W[k]$ for $\forall k$. then the sum of this product would be calculated and the superpixel having the maximum number would be considered as the marker for the that frame. This can be formalized as follow:
\begin{equation}
S_{[i, m, n]} = \sum_{\forall k}(N_{k[i, m, n]} \times W[k])
\end{equation}
\begin{equation}
D_{[m,n]} = S_{[Ind(max(S_{[\forall i, m, n]})), m, n]}
\end{equation}
where $S_{[i,,m,n]}$ is the sum of weighted features of the marker $m$ calculated for the frame $n$. $Ind()$ finds the index of $S_{[i,,m,n]}$ which is equal by $max(S_{[i,,m,n]})$. The process for tracking is illustrated in Figure \ref{fig4}.

\section{Results} \label{Results}
The method was examined using Python 2.7.12 platform with installed OpenCV 3.1.0-dev on a MacBook pro 2.7 GHz Intel Core i5 with 8 GB 1867 MHz DDR3.

\begin{table*}[t]
\caption{Tracking results. The "SLIC + 3D Tracking" is the method presented here which is compared with three methods; "SLIC + 2D Tracking" \cite{maghsoudi2017superpixels}, "Thre + 2D Tracking" \cite{maghsoudi2017superpixels}, and "Manual Tracking". "SLIC" and "Thre" are superpixel method presented in \cite{Achanta12} and the thresholding on the hue channel. There are eight small tables showing different conditions for evaluation of the method. From top to bottom the small tables show: the number of frame and markers for the whole database used here; average time was required for each method to process one trial (1000 frames); the results for tracking of the markers drawn bad or unclear, the tracking results when the markers were unclear or hidden from the beginning; the tracking results while a marker was partially occluded; the results when the markers were completely occluded; the results when a perfect consequence of frames are next to each other; the comprehensive results considering all mistakes and conditions excluding the "Missing Start". } 
\label{tab:fonts} \label{tab1}
\begin{center}    
\begin{tabular}{ | c | c | c | c | c | c |}
    \hline
    Method & Manual Tracking & Thre + 2D Tracking & SLIC + 2D Tracking & SLIC + 3D Tracking\\ \hline
    \textbf{Database Frames} & 1,000 & 4,000 & 4,000 & 24,000\\ \hline
    Database Markers & 5,000 & 20,000 & 20,000 & 120,000\\ \hline  \hline
    \textbf{Average Time per Trial} & \textbf{$8,700 \pm 2300$} & \textbf{$108 \pm 12$} & \textbf{$118 \pm 14$} & \textbf{$149 \pm 18$}\\ \hline \hline
    
    \textbf{Bad Marker Frames} & - & 800 & 800 & 13,500\\ \hline
    Total Bad Markers & - & 800 & 800 & 13,500\\ \hline
    Correct Tracked & - & 23 & 127 & 11,582\\ \hline
    Percentage & \textbf{100} & \textbf{2.88} & \textbf{15.87} & \textbf{85.79}\\ \hline \hline
    
    \textbf{Missing Start} & 100 & 400 & 400 & 2,200\\ \hline
    Total Markers & 500 & 2,000 & 2,000 & 11,000\\ \hline
    Correct Tracked & 500 & 35 & 59 & 1319\\ \hline
    Percentage & \textbf{100} & \textbf{1.75} & \textbf{2.95} & \textbf{11.99}\\ \hline \hline
    
    \textbf{Partitially Occluded} & 30 & 250 & 250 & 4,500\\ \hline
    Total Markers & 150 & 1,250 & 1,250 & 22,500\\ \hline
    Correct Tracked & 150 & 10 & 217 & 21,212\\ \hline
    Percentage & \textbf{100} & \textbf{0.8} & \textbf{17.36} & \textbf{94.28}\\ \hline \hline
    
    \textbf{Occlused} & 50 & 300 & 300 & 5,100\\ \hline
    Total Markers & 250 & 1,500 & 1,500 & 25,500\\ \hline
    Correct Tracked & 250 & 10 & 217 & 22,788\\ \hline
    Percentage & \textbf{100} & \textbf{0.67} & \textbf{14.47} & \textbf{89.36}\\ \hline \hline
    
    \textbf{Perfect Consecutive} & 450 & 1,300 & 1,300 & 8,100\\ \hline
    Total Markers & 2,250 & 6,500 & 6,500 & 40,500\\ \hline
    Correct Tracked & 2,250 & 6,377 & 6,494 & 40,496\\ \hline
    Percentage & \textbf{100} & \textbf{98.11} & \textbf{99.90} & \textbf{99.99}\\ \hline \hline
    
    \textbf{Total Frames} & 900 & 3,600 & 3,600 & 21,800\\ \hline
    Total Markers & 4,500 & 18,000 & 18,000 & 109,000\\ \hline
    Correct Tracked & 4,500 & 10,891 & 14,237 & 103,562\\ \hline
    Percentage & \textbf{100} & \textbf{60.50} & \textbf{79.09} & \textbf{95.01}\\ \hline
\end{tabular}
\end{center}
\end{table*}

Applying SLIC superpixel method on the frame or the generated sub-images, to reduce the required time for the superpixel process \cite{Achanta12, maghsoudi2017superpixels}, was the segmentation process as illustrated in Figure \ref{fig4}. As shown in Figure \ref{fig1}, the number of superpixels plays an important role in how the SLIC method would be performed. We had a comprehensive discussion on how we can select the correct superpixel number based on the size of markers \cite{maghsoudi2017superpixels}. If the size of marker would be a known parameter, then, the following equation can find the best superpixel number (NSLIC):
\begin{equation} \label{eq13}
NSLIC = \dfrac{2048 \times 700}{NP} \times  \dfrac{1}{100 \times 100 \times 2}
\end{equation}
where $NP$ is the number of pixels for that marker. 2048, 700, and 100 are the image width, height, and sub-image window size, respectively. Equation \ref{eq13} can provide an ideal number of superpixels for SLIC; however, we needed an estimation of the size as the SLIC can segment the objects with half up to twice of initial size. Therefore, we considered 10,000, 10,000, 7,000, 3,000, and 3,000 as a number of superpixels of a frame, $NSLIC$, for toe, ankle, knee, hip, and anterior superior iliac spine markers, respectively.

The segmentation process using the SLIC superpixel method was examined in \cite{maghsoudi2017superpixels}. 

As discussed in the introduction, manual tracking can be considered as the common method to track the markers for many applications in biomechanics. To compare how the proposed method can be helpful for biomechanics/neuroscience applications; we compare this method with manual tracking, thresholding for segmentation and 2D tracking, SLIC method for segmentation and 2D tracking \cite{maghsoudi2017superpixels}, SLIC method for segmentation and 3D tracking. 

\begin{figure*}[t!p]
\centering
\includegraphics[width=0.8\textwidth]{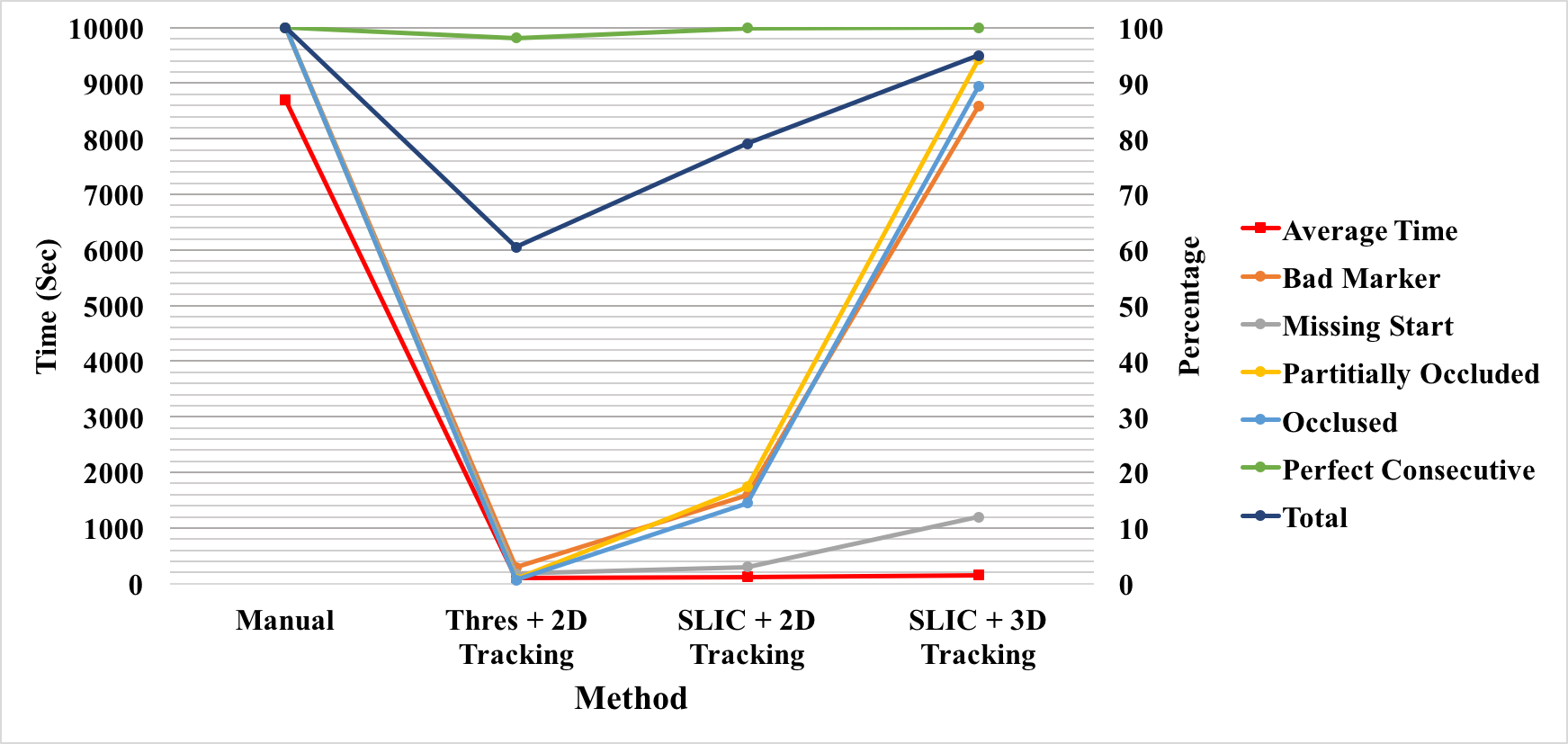}
\caption{A comparison between the methods. Manual tracking, thresholding following by 2D tracking, SLIC following by 2D tracking, and the method presented here are compared with each other. The Average time to process a trial, 1,000 frames, is graphed in red. The average time is on the left axis and the rest of plots are on the right axis.} \label{fig5}
\end{figure*}

The method was examined in six Sprague-Dawley rats. Each rat had five markers showing: toe, ankle, knee, hip, and anterior superior iliac spine. We randomly selected two trails from each rat and each trial contained 1,000 frames. It created 12,000 from each of the two cameras capturing the right side of the animal, the five markers were drawn on the right side. Therefore, total 24,000 frames producing 120,000 markers consist of this study database.

We evaluated the method for different conditions: bad marker frames, the marker was painted poorly causing difficulties in finding them; missing start, a trial starts with a set of markers barely being visualized but still user initialized the marker location by guessing the position; partially occluded, the markers were partially occluded by body or dirt on the plexy glass; occluded, the markers were completely occluded; perfect consecutive, a consequence of frames that the markers were clear for whole time; total frames, the total results were reported. 

\begin{figure*}[t!p]
\centering
\includegraphics[width=0.6\textwidth]{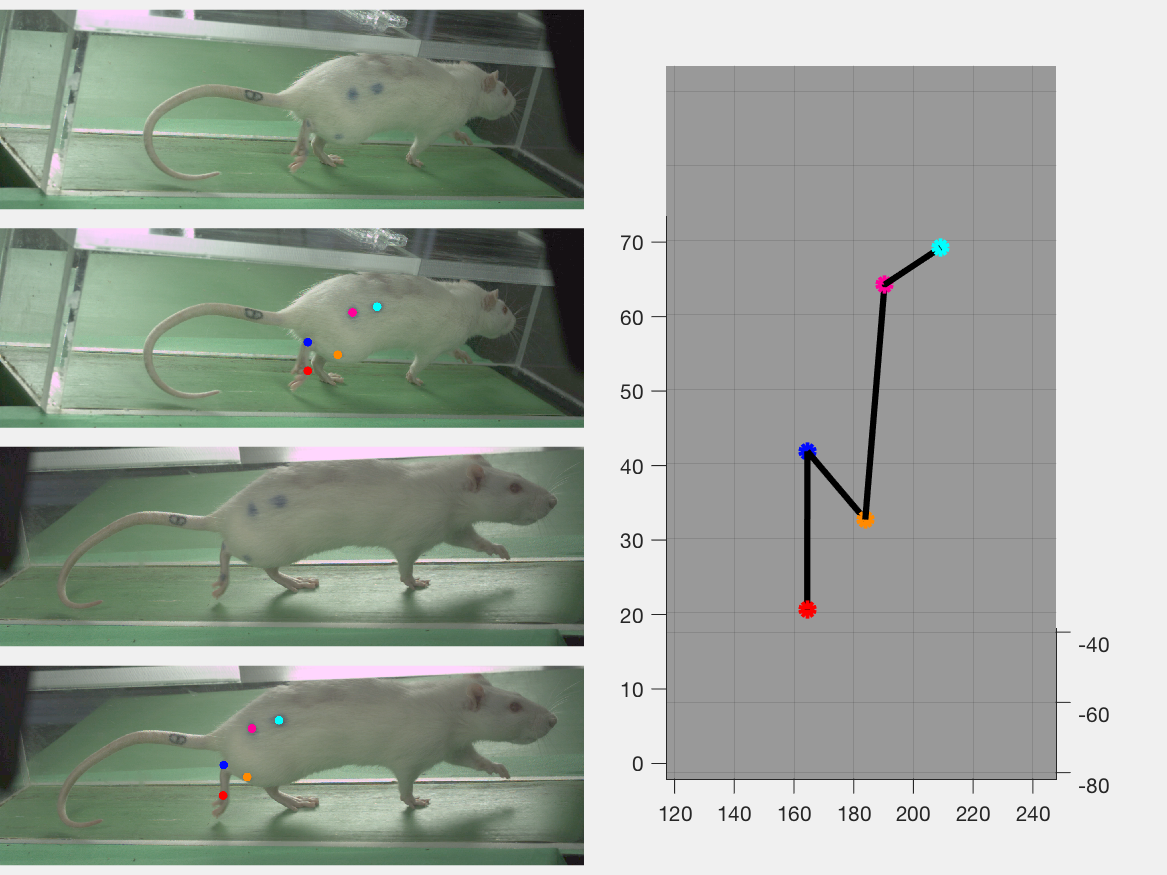}
\caption{A video of tracked markers for 1,000 consecutive frames using the presented method here. (MP4 14.6MB). Left frames from top to bottom show a captured frame from camera 3, tracked markers for the corresponding frame from camera 3, a captured frame from camera 4, and tracked markers for the corresponding frame from camera 4. The right image shows the 3D reconstruction of markers using the DLT coefficients. } \label{fig6}
\end{figure*}

The results for all these conditions are separately illustrated in Table \ref{tab1} and Figure \ref{fig5}. In addition, Figure \ref{fig6} shows a 3D reconstructed video from a rat while running on the treadmill. 

It should be noted that we did not add the required time for finding the DLT coefficients in the results presented here.

\section{Conclusion}

We presented a method to segment the drawn markers on the body of rats using SLIC superpixel method following by a 3D Kalman based tracker to predict the position of markers in a 3D domain and projecting them to the 2D image plane. Having the coordinates on the 2D image plane and assigning a score to each of the superpixels based on the predicted coordinate, color, and texture information of marker in the previous frame provided us the ability to use a probabilistic function \ref{fig4}. 

The method was evaluated 24 trails and 5 markers drawn on the body of an animal.  We compared the method with available methods \cite{maghsoudi2017superpixels} utilizing simple thresholding or superpixel method followed by 2D tracker based.

The results showed that the best method, as expected, was using manual tracking; however, it takes so much time to process one trail. It shows the importance of using an accurate method for marker tracking. In addition, the manual tracking involves intraobserver and interobserver tracking error which was not studied here. 

The 3D tracking showed its superiority compared with 2D tracking methods in all conditions. However, the results show that if there would be a perfect consequence of frames, the superpixel method using 2D tracking can work the same as 3D based tracking method while it is slightly faster than 3D based tracking method. It should be reminded that the required time to calculate the DLT coefficients was not involved in the time plot in Table \ref{tab1}. However, it is hard to find perfect consequence of frames when capturing from the animal.

Our future study would be developing 3D based tracker for a markerless animal to avoid handling and painting of markers on the body of the animal. The painting a marker needs long anesthesia for mice following by bleaching and drawing markers. In addition, we will try to use a 3D model of markers to reduce the miss-tracking. Having a 3D model can provide a good setup to track the markers based on the other markers. It means that we can find the markers missing or wrongly labeled using the other markers.

{\small
\bibliographystyle{ieee}
\bibliography{report}
}

\end{document}